\documentclass[letterpaper, 10 pt, conference]{ieeeconf}  
\IEEEoverridecommandlockouts                              
\usepackage{graphicx}
\usepackage{subfig}
\graphicspath{{images/}}
\usepackage{times}    
\usepackage{svg}
\usepackage{amsmath}     
\usepackage{amssymb}     
\usepackage{graphicx}
\usepackage[noend]{algpseudocode}
\usepackage{url}
\usepackage{amsmath}
\usepackage[linesnumbered,ruled]{algorithm2e}
\usepackage{siunitx}
\usepackage{colortbl}
\let\labelindent\relax
\usepackage{enumitem}
\usepackage{cite}

\usepackage[font=small]{caption}

\def\secref#1{Sec.~\ref{#1}}
\def\figref#1{Fig.~\ref{#1}}
\def\tabref#1{Tab.~\ref{#1}}
\def\eqref#1{Eq.~(\ref{#1})}

\newcommand\etal{\emph{et al. }}

\newcommand*{\algotitle}[2]{%
  \stepcounter{algocf}%
  \hypertarget{algocf.title.\theHalgocf}{}%
  \NR@gettitle{#1}%
  \label{#2}%
  \addtocounter{algocf}{-1}%
}

\title{\LARGE \bf Online Object-Oriented Semantic Mapping and Map
  Updating}

\author{Nils Dengler \and Tobias Zaenker \and Francesco Verdoja \and Maren Bennewitz
\thanks{N. Dengler, T. Zaenker, and M. Bennewitz are with the Humanoid
  Robots Lab, University of Bonn,  Germany.  F.  Verdoja  is with
  School of Electrical Engineering, Aalto University, Finland. This
  work has been partially funded by the Deutsche
  Forschungsgemeinschaft (DFG, German Research Foundation) under
  Germany's Excellence Strategy -- EXC 2070 -- 390732324 (PhenoRob)
  and by the Academy of Finland Strategic Research Council grant 314180.\newline
  978-1-6654-1213-1/21/\$31.00~\copyright~2021 IEEE
  }
  }

\begin{document}
\maketitle
\thispagestyle{empty} 
\pagestyle{empty}

%

\begin{abstract} Creating and maintaining an accurate representation of the environment is an essential capability for every service robot.
 Especially for household robots acting in indoor environments, semantic information is important. 
 In this paper, we present a semantic mapping framework with modular map representations. 
 Our system is capable of online mapping and object updating given object detections from \mbox{RGB-D}~data and provides various 2D and 3D~representations of the mapped objects. 
 To undo wrong data associations, we perform a refinement step when updating object shapes. 
 Furthermore, we maintain an existence likelihood for each object to deal with false positive and false negative detections and keep the map updated. 
 Our mapping system is highly efficient and achieves a run time of more than 10 Hz. 
 We evaluated our approach in various environments using two different robots, i.e., a Toyota HSR and a Fraunhofer \mbox{Care-O-Bot-4}. 
 As the experimental results demonstrate, our system is able to generate maps that are close to the ground truth and outperforms an existing approach in terms of intersection over union, different distance metrics, and the number of correct object mappings. 
\end{abstract} 


\section{Introduction}
\label{sec:intro} 
In any mobile robotics application, maps are of great importance.
While geometrical map representations such as grid maps are an
established researched topic, the research on mapping also semantic information becomes more and more popular.
Especially for service robots, a proper environment understanding is necessary to not only navigate but also interact reasonably with the world.
To generate accurate semantic maps for navigation, most research concentrates on object-based SLAM algorithms \cite{slam} or the complete 3D reconstruction of the world with semantic annotations \cite{kimera}.
However, the interest of maps with focus on just the objects has increased noticeably\mbox{\cite{meaningful, Li, zaenker, efficient, voxx}}. 
One of the main reasons is the usefulness of concrete object instances, which can be accessed more easily in an object-based approach.
Most current approaches that create object-oriented maps have in common that they provide a 3D object reconstruction together with the position in the map.
While the 3D information is necessary for any kind of manipulation,
for navigation tasks often 2D~information is sufficient~\cite{regier2020classifying}.

\begin{figure}
\centering
\includegraphics[scale=0.45]{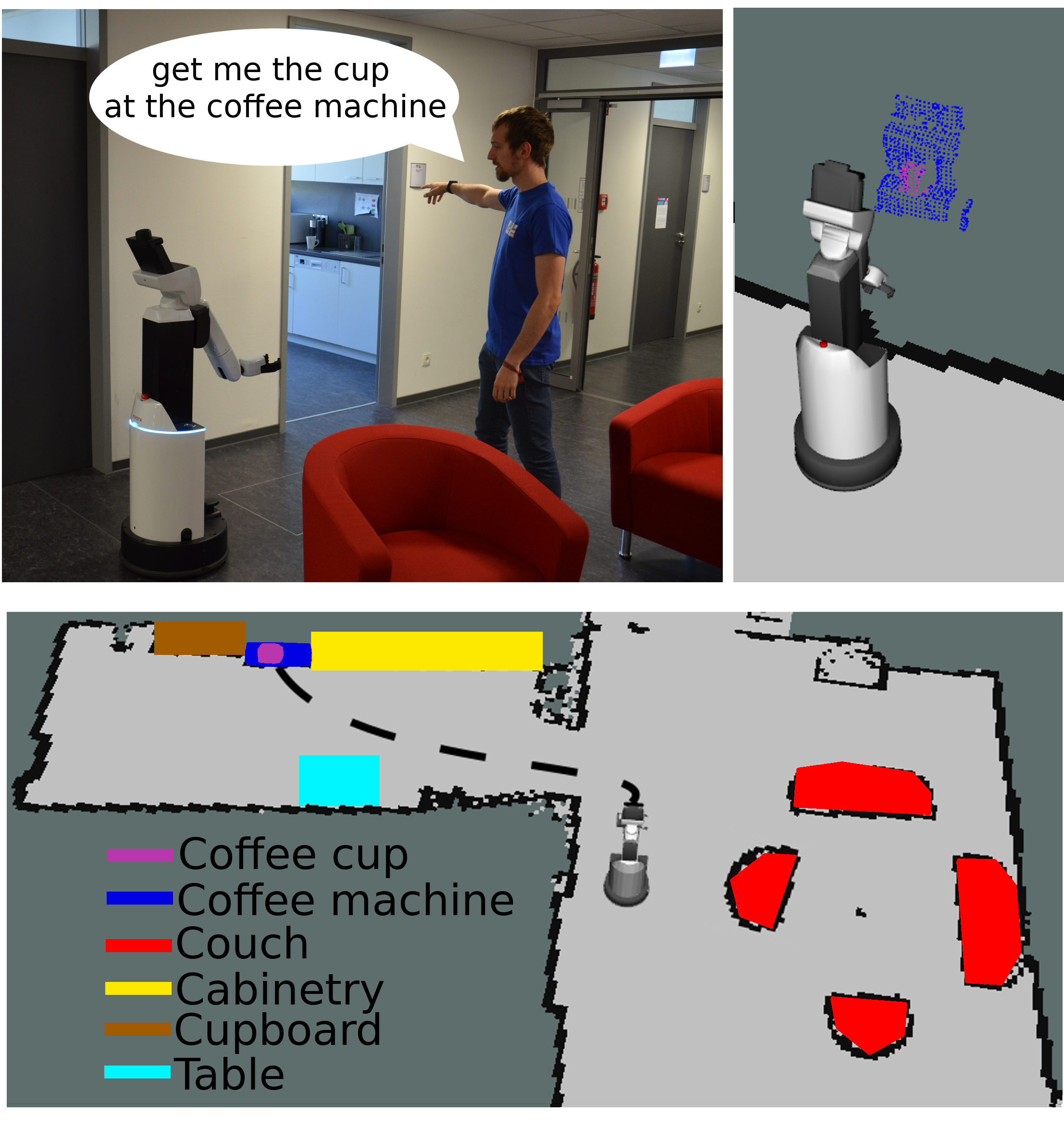}
\caption{Example usecase of our proposed system. A user commands the robot to get a specific cup~(upper left image).
To get the position of the cup, the robot checks the created map for cups near
a coffee machine. The lower image shows a 2D~polygonal representation
of the environment mapped with
our framework. 
With the 2D~information, the robot can navigate to the goal destination.
To grab the cup, the robot then utilizes the 3D representation of the
objects~(upper right image), which is also maintained by our system.}
\label{cover}
\end{figure}

In our approach, we maintain for each object instance multiple object
information to address different applications.  
Apart from the 3D point cloud describing the object and its semantic type, also the 2D
shape in form of a polygon and an oriented bounding box are stored.
Due to the modularity, it is easily possible to extend our system with additional task-related representations.
A possible application of our system is shown in \figref{cover}.  
With the mapped representations of objects in the environment, the
robot is able to identify the requested cup by checking for all cups close to a coffee machine.
With this information, the robot plans a path to the cup according to the mapped 2D shapes.
For grabbing the cup, the 2D shape is not sufficient anymore and the robot can use the 3D point cloud representation.

For the mapping process, we assume the global robot pose to be known.  
As in other approaches, we combine instance-level segmentation with a fast geometric segmentation to reconstruct objects
from the incoming \mbox{RGB-D}~data.
To be robust against dynamic changes in the environment, we developed a data association that is able to update mapped objects with newly collected object information.  
Our data association is hereby able to handle wrong object updates due to false segmentation.
While existing mapping approaches update only currently observed objects~\cite{meaningful}, we introduce an object-wise likelihood that is also updated for mapped objects not detected in the current field of view.

To sum up, in this paper we present a novel modular semantic mapping system with the following contributions:
\begin{itemize} 
\item A map representation that provides several object information such as the object label, object point cloud, and two different 2D object shapes,
\item A robust data association that is able to undo wrong object updates,
\item An existence likelihood calculation for each mapped object to deal with false detections and keep the map updated,
\item An experimental evaluation of our system in three different environments that shows the qualitative performance as well as the superior performance in comparison to another semantic mapping approach by Zaenker~\etal \cite{zaenker} with respect to different metrics.
\end{itemize} 


\begin{figure*} \centering 
\includegraphics[width=\textwidth]{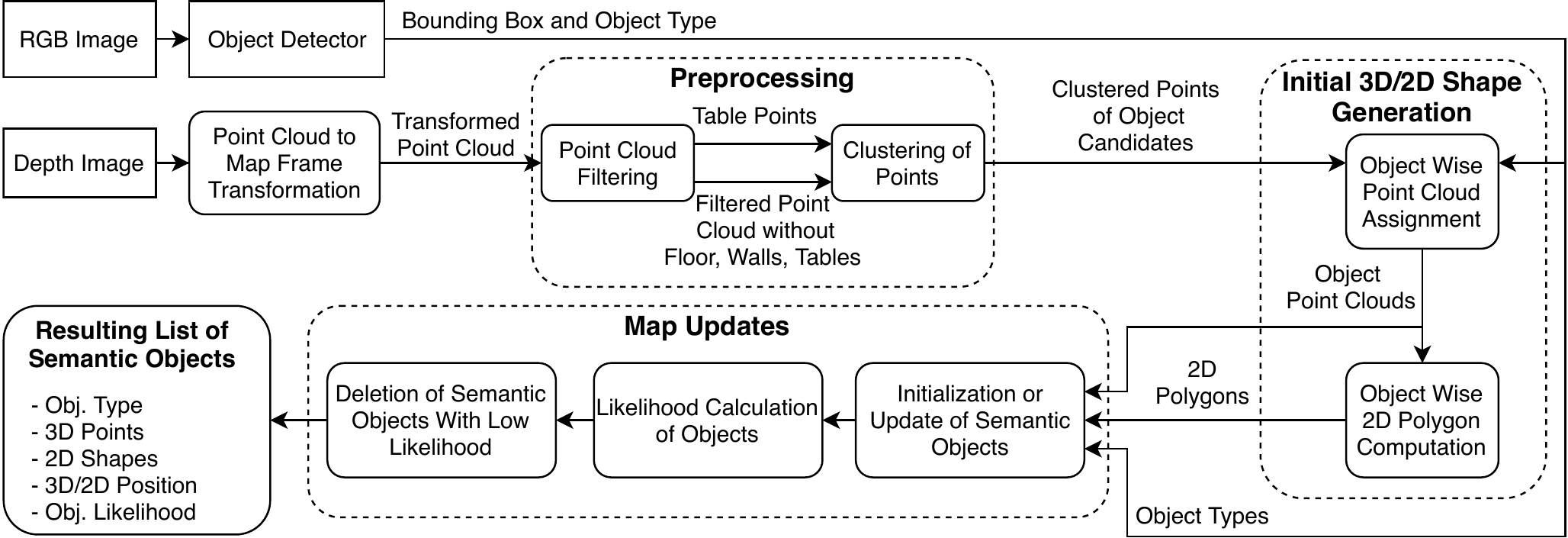} \caption{Overview of our semantic mapping system.
The squared boxes indicate the external inputs.} 
\label{flow}
\end{figure*} \begin{figure*}[] \centering 
\subfloat[Original point cloud\label{preproc_res:a}]{\includegraphics[height=1.5in]{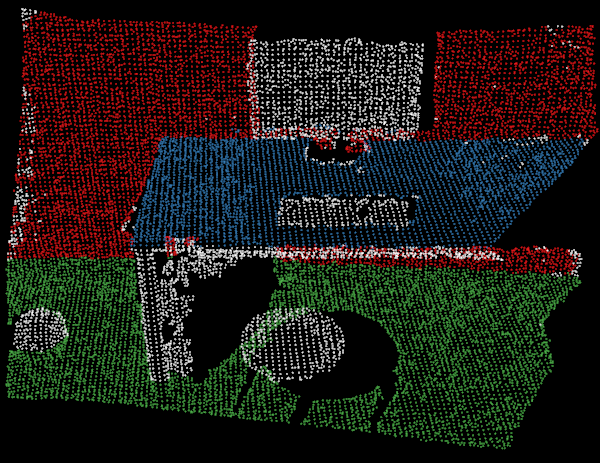}} \hspace{0.5em} \subfloat[Point cloud representation \label{preproc_res:b}]{\includegraphics[height=1.5in]{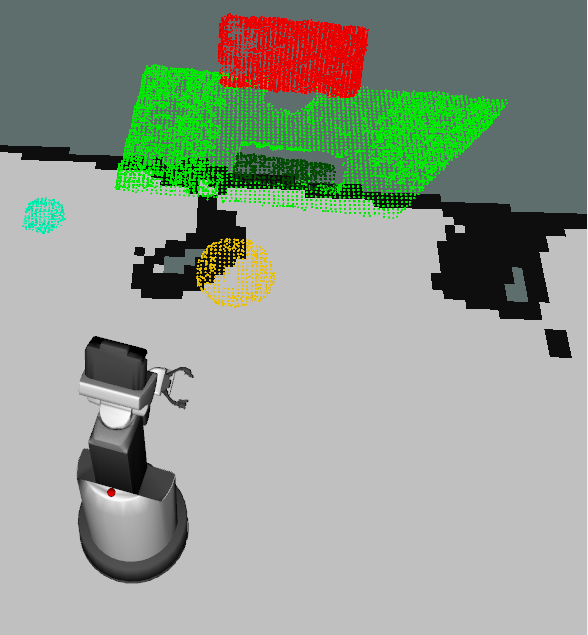}} \hspace{0.5em} \subfloat[Polygon representation \label{preproc_res:c}]{\includegraphics[height=1.5in]{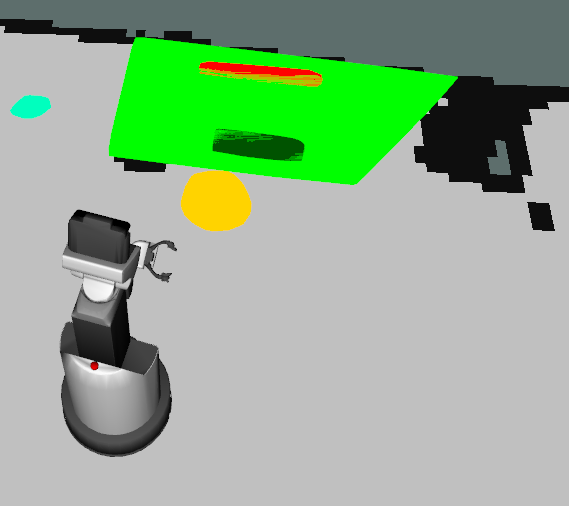}} \hspace{0.5em} \subfloat[OBB representation \label{preproc_res:d}]{\includegraphics[height=1.5in]{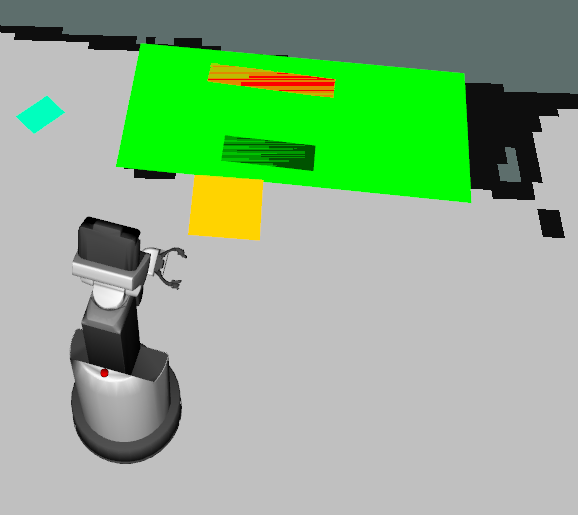}} \caption{Office working place, observed by the robot.
(a) visualizes the whole point cloud with the three filtered surfaces colored in red (wall), blue (table) and green (floor).
All remaining parts of the point cloud after the preprocessing are colored white.
Figures (b) to (d) show the mapped objects in three different representations (point cloud, polygon, and oriented bounding box (OBB)).} 
\label{preproc_res}
\end{figure*}

\section{Related Work}
\label{sec:related}



For many scenarios, semantic 2D maps are sufficient, e.g., for simple search and navigation tasks.
For example, \mbox{Regier~\etal \cite{regier2020classifying}} proposed
to label cells of a 2D gridmap with object information and corresponding navigation cost to overcome the obstacles.
Leidner~\etal \cite{leidner2014object} presented a semantic mapping approach to augment existing geometric 2D maps.
The authors demonstrated the automatic annotation of sub spaces on the example of warehouse environments.
The main difference to our approach is that these frameworks do not output single object instances, but only the labeled geometric 2D map.

The research on object-oriented semantic mapping has increased in the
last few years.
The five approaches closest to our work are \cite{meaningful, Li, zaenker, efficient, voxx}, which all address the problem of providing concrete object information.
Object-oriented mapping approaches can be generally separated into two main parts, the applied object instance segmentation and the geometric segmentation.
For the object instance segmentation, two variants are mainly used.
In several approaches, \mbox{mask R-CNNs}~\cite{he2017mask} are
applied to get an accurate instance mask of the detected objects~\cite{Li, zaenker, voxx}.
Other approaches use detectors that provide only bounding boxes~\cite{meaningful,efficient}.
The most popular ones are object detectors such as YOLO \cite{yolov2}, the TensorFlow object detection API \cite{rostf}, and single shot detectors \cite{ssd}, which provide many pretrained models that can be used off-the-shelf.
For this reason, we use a pretrained \mbox{faster R-CNN} \cite{ren2015faster}  from TensorFlow~\cite{rostf}.
Note that it can be replaced by, e.g., a \mbox{mask R-CNN} or the faster \mbox{Yolact++} \cite{yolact++} with just minor changes.\\ 
\indent For the geometric segmentation, the research towards supervised methods has increased \cite{rosu2020semi}\cite{stekovic2020casting}.
However, those approaches need prior knowledge and training data.
Our system, in contrast, does not need any prior knowledge about the object shape.
In most approaches, the objects are already segmented in the depth image before computing the corresponding 3D points~\cite{meaningful,efficient,voxx}. 
While there is no apparent benefit in computing the segments on the depth image and the point cloud has to be computed anyway, we carry out the segmentation on the point cloud.
With a cluster computation time of less than 1\,ms, our approach is faster than, e.g., S{\"u}nderhauf~\etal\cite{meaningful}.\\
\indent A different segmentation strategy is proposed by Li~\etal\cite{Li}. Instead of applying geometric segmentation the authors use a Gaussian mixture model to refine the mapped object regions generated according to the segmentation masks.\\
\indent While these approaches also generate object-oriented semantic maps as our system, there are three major benefits that are not addressed by the other works.
All approaches have in common that the represent the world solely in 3D.
Since the map representation depends heavily on the task, our approach is not limited to just a 2D or 3D representation, but combines multiple variants in a modular way.
A further benefit is that our approach performs a more robust data association and can fix possible wrong object matchings.
Finally, we maintain an existence likelihood for each object.
Typically, object-oriented approaches update the objects only if they
are re-detected~(e.g., \cite{meaningful}).
That means that especially if the environment is non-static and objects change their location, the probability of misinterpreting the new object points as the old ones is quite high.
In contrast, our proposed system is able to manage these changes and
update the map even if objects are no longer present at their previous
location.

Our semantic mapping framework follows ideas of the
hypermap framework~\cite{zaenker} but has several novel
contributions. Hypermap is also an object-oriented semantic mapping
system, however, in contrast to our approach, it only maintains a 2D representation.
We provide and maintain object shapes in 3D and developed a novel method for data
association and dealing with wrong object updates.

\section{Object-Oriented Semantic Mapping}
\label{Object_shape}
In this section, we describe the four main parts of our system.
The semantic map generated by our approach consists of $N$ semantic
objects $o_1, \dots, o_N$, which are stored in a list.
Each $o_i$ is defined by several features such as the object type, the
corresponding point cloud, the 2D shape polygon and oriented bounding
box~(OBB), as well as the 2D and 3D position, and its existence likelihood.
The first part of our framework is the preprocessing and segmentation of the point cloud.
Then each segmented part is assigned to a detected object in the RGB~image.
Subsequently, our system checks whether the object can be associated to an already mapped semantic object, if not, the new objects is included into the semantic map.
If an already existing semantic object is updated, the likelihood is
re-calculated and used to remove or add objects.
Fig.
\ref{flow} shows an overview of our system.
The individual steps are described in detail in the following.

\subsection{Detection of Objects}
\label{Detection}

In semantic mapping, the interest in using instance segmentation masks, e.g., \mbox{R-CNNs}~\cite{he2017mask} as object detector to get more precise object instances, has increased~\cite{Li,voxx}.
However, a major restriction of current object detectors is the limited amount of classes in freely available pretrained nets.
While the popular COCO dataset \cite{Coco} can detect up to 80 different classes, important objects for office environments such as tables are missing.
Therefore, we use a faster \mbox{R-CNN} trained on the OpenImages-Dataset~\cite{oid}.
This dataset includes more than 600 classes, with almost any object commonly seen in households or offices.
Note that our framework is modular and the detection method is easily replaceable.
Note then even when  instance masks are used, clustering in the point cloud~(cf.~\secref{Filtering})
is still necessary since the instances are imperfect.

\subsection{Point Cloud Preprocessing}
\label{Filtering}

In a first preprocessing step, our system removes the floor and the walls.
To facilitate the segmentation of smaller objects on tabletops, they are also removed from the initial point cloud but the corresponding points are stored separately since tables are important semantic objects.
We hereby apply the RANSAC algorithm
\cite{fischler1981random}. Fig.~\ref{preproc_res:a} shows the result
of the filtering step for an example scene.
Furthermore, we perform a statistical outlier removal step to increase the accuracy of the point cloud and counter the assignment of wrong points before the geometric segmentation.

\subsection{Geometric Segmentation}

As next step, we apply geometric segmentation on the preprocessed point cloud.
To perform the geometric segmentation, we apply the Euclidean cluster extraction of PCL~\cite{Rusu_ICRA2011_PCL}.
We compute the clusters for the whole preprocessed point cloud instead
of just the points corresponding to the  objects, since
more continuous information is beneficial for the clustering.
Because there is a high probability to also have very small objects in the environment, e.g., cups or phones, we hereby use a minimum cluster size of 100 points, that depends on the point cloud density.
After the segmentation, we store the generated clusters to combine them with the bounding boxes of the object detector.

\subsection{Initial 3D and 2D Object Shape Generation}
\label{2D_Shape_generation}

Given the bounding box of each object and the generated clusters, our system assigns the best fitting cluster to each bounding box.
To do so, it first determines the cluster that has the highest number of points inside the object bounding box.
Afterwards, the corresponding points inside the box are stored as the object point cloud.
To generate the initial 2D~object shape, all 3D points of the object point cloud are projected onto the 2D $x$-$y$-plane and the convex hull of the remaining points is computed to express it as a polygon.
As further 2D~shape variant, we also compute the oriented bounding box~(OBB) using the rotating calipers method~\cite{calipers}.
In contrast to the polygon, the OBB can be a better approximation of the real shape if parts of an object are obscured.
Fig.~\ref{preproc_res:b}-d show the representations of part of an office scene.

Furthermore, we determine the 2D and 3D~position as the center of mass (CoM) of the corresponding representation.



\subsection{Initialization and Update of Semantic Objects}
\label{Combination_of_shapes}

%

\begin{figure}
\centering
\subfloat[with refinement]{\includegraphics[height=1.4in]{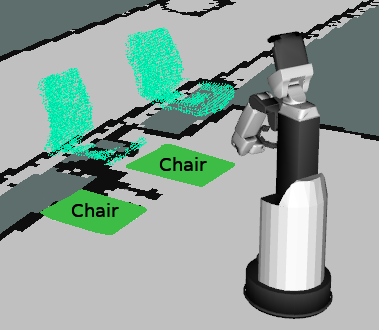}}
\hspace{0.5em}
\subfloat[without refinement]{\includegraphics[height=1.4in]{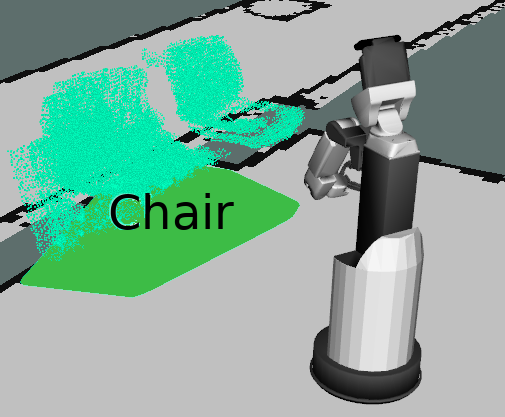}}
\caption{Example of the mapping part of the environment (a)~with
  and (b)~without object shape refinement during updates. Shown are
  the object point cloud, the 2D polygons and the label. In (a) the two chairs are clearly separated, while in (b) they are merged together due to many outlieres resulting from, e.g., wrong segmentation or localization unaccuracy. }\label{chair_sep}
\end{figure}

Data association is a key element of every semantic mapping framework.
To determine whether to insert a new object or update an existing $o_i$ when processing a new detection, our system uses an R-tree structure \cite{rtree} to find overlapping polygons of the same object type, which can be combined.
The update process then consists of one of the following two cases.

\begin{description}[font=\normalfont\itshape] 
\item[New object:]
In the first case, the initial polygon has no intersection with any already mapped object and therefore our system creates a new semantic object.
\item[Association with at least one existing semantic object:] In the
second case, the point clouds of all overlapping semantic objects are combined together with the newly found object to create a single semantic object.
We hereby filter the combined point cloud with a voxel grid to get evenly distributed points.
As described in Section \ref{2D_Shape_generation}, the polygon of the corresponding combined objecs is updated as well as the OBB.
While all previously combined objects are deleted from the R-tree, the updated object is newly included.
\end{description} 

It is possible that in the update process wrong points are associated to an object, due to errors in the segmentation process or localization inaccuracies.
In this case, there is a high probability that close-by objects of the same class are combined because of the expanded object shapes.
To refine the object shape and ensure that multiple incorrectly combined objects can be separated again, our system calculates new Euclidean clusters of the object point cloud if an $o_i$ was combined more than a predefined amount of times.
If the algorithm finds more than one cluster in the point cloud of an $o_i$, the biggest cluster is assigned to the $o_i$ and new semantic objects are created for the other clusters.
Depending on the likelihood parameters the system needs only few
further observations to undo wrong mappings that could arise.
Fig. \ref{chair_sep} shows the mapping of two chairs with and without
this object refinement step.
Without the object refinement, points from other structures are included in the point cloud, which wrongly extends the shape.
Furthermore, points of the other chair intersect with the first one and lead to the combination of the two objects.
With our refinement step, the objects are separated and have sharp boundaries.

\subsection{Object Likelihood}
\label{Likelihood_calculation}

To keep the map updated in case of object displacements or removals as well as to deal with false detections, object deletion is an important property of our system.
In other frameworks such as \cite{meaningful}, the map is only updated wrt.
the semantics. If the object is re-observed and in case an object is removed or displaced by another object, the semantic object instance remains in the map.
To counter this problem, we calculate a likelihood for the existence
of each semantic object each time the object should be in the field of
view~(FoV), i.e., our system counts the number of times an object was
initialized or updated  (hit) or not (miss).
After each iteration, our system updates the likelihood of object~$i$ as follows: 
\begin{align}\label{counting}
 L_i = \alpha_{it} \cdot \frac{hits_i}{\frac{1}{hit_i+misses_i}+misses_i + hits_i} 
\end{align}
The higher the likelihood, the higher the confidence that the object really exists.
With the term $\frac{1}{hit_i+misses_i}$ we achieve a smooth likelihood evolution and keep some uncertainty for the objects. 
$\alpha \in [0,1]$ indicates the survival time factor of each object and can be calculated as suggested in \cite{survival}. 
In our experiments, we used $\alpha_{it} = 1$, which means that we do not
incorporate any prior knowledge of object dynamics.
To delete objects from the map, we use a threshold $\tau$.
If $L_i < \tau$, the corresponding semantic object will be deleted.
As a constraint, the summed up maximum amount of hits and misses is limited to a predefined number so that it cannot grow infinitely and detections far in the past are not weighted.
To accomplish this, the oldest detection is deleted and the newest one
is added. 
Furthermore, we consider a minimum number of summed up hits and misses to encounter the inconsistency of the object detector and ensure that early false negative detections will not affect the mapping, such that the object is removed although it is still in the environment.
For our experiments, we defined a maximum number of 30 observations, a minimum
number of 10, and a deletion threshold of~0.25.



\section{Experimental Evaluation}
\label{sec:exp}
We evaluated our approach on a computer with an \mbox{i9-9900K} eight-core CPU at 3.60 GHz and an Nvidia 2080 GPU with \mbox{8 GB} of memory used for the faster R-CNN.
Our approach assumes the robot pose to be given and applies Monte Carlo
localization~\cite{fox1999monte} on a map generated with $gmapping$~\cite{gmapping}.
In the following, we first evaluate the online mapping capability of our system in different environments and compare it to the closely related  approach by Zaenker \etal \cite{zaenker}, which is freely available.
Then we demonstrate the evolution of the object likelihood values and,
finally, analyze the runtime of our system.

\subsection{Online Object Mapping and Map Updates}
\begin{table*} \centering 
\resizebox{.49\textwidth}{!}{%
\begin{tabular}{|c|c|c|c|c|c|} \hline \rowcolor[gray]{.75} \textbf{Office 1} & \textbf{ IoU$\ast$} & \textbf{H-Dist$\ast$}&\textbf{CoM-Dist$\ast$} & \textbf{TP}& \textbf{FP} \\ 
Polygon & 0.58 $\pm$ 0.16& 0.12m $\pm$ 0.05& 0.04m $\pm$ 0.02& 20/23&4 \\ \hline Obb & 0.51 $\pm$ 0.18& 0.14m $\pm$ 0.06& 0.04m $\pm$ 0.02& 20/23&4 \\ \hline H-map \cite{zaenker}& 0.35 $\pm$ 0.18& 0.44m $\pm$ 0.24& 0.21m $\pm$ 0.11&16/23&1 \\ \hline \end{tabular}} 
\resizebox{.49\textwidth}{!}{%
\begin{tabular}{|c|c|c|c|c|c|} \hline \rowcolor[gray]{.75} \textbf{Office 2} & \textbf{ IoU$\ast$} & \textbf{H-Dist$\ast$}&\textbf{CoM-Dist$\ast$} & \textbf{TP}& \textbf{FP} \\ 
Polygon & 0.51 $\pm$ 0.18& 0.17m $\pm$ 0.12& 0.08m $\pm$ 0.05& 21/23&10 \\ \hline Obb & 0.45 $\pm$ 0.18& 0.18m $\pm$ 0.13& 0.08m $\pm$ 0.06& 21/23&10 \\ \hline H-map \cite{zaenker}& 0.29 $\pm$ 0.15& 0.55m $\pm$ 0.44& 0.27m $\pm$ 0.23&18/23&8 \\ \hline \end{tabular}}\\ \vspace{.5em} 
\resizebox{.49\textwidth}{!}{%
\begin{tabular}{|c|c|c|c|c|c|} \hline \rowcolor[gray]{.75} \textbf{Large} & \textbf{ IoU$\ast$} & \textbf{H-Dist$\ast$}&\textbf{CoM-Dist$\ast$} & \textbf{TP}& \textbf{FP} \\ 
Polygon & 0.49 $\pm$ 0.15& 0.24m $\pm$ 0.15& 0.09m $\pm$ 0.06& 25/33&8 \\ \hline Obb & 0.44 $\pm$ 0.16& 0.27m $\pm$ 0.16& 0.10m $\pm$ 0.08& 25/33&8 \\ \hline H-map \cite{zaenker}& 0.26 $\pm$ 0.12& 0.53m $\pm$ 0.17& 0.25m $\pm$ 0.10&18/34&1 \\ \hline \end{tabular}} 
\resizebox{.49\textwidth}{!}{%
\begin{tabular}{|c|c|c|c|c|c|} \hline \rowcolor[gray]{.75} \textbf{Cluttered} & \textbf{IoU} & \textbf{H-Dist}&\textbf{CoM-Dist} & \textbf{TP}& \textbf{FP} \\ 
Polygon & 0.29 $\pm$ 0.19& 0.45m $\pm$ 0.30& 0.21m $\pm$ 0.21& 19/33&8
\\ \hline Obb & 0.29 $\pm$ 0.21& 0.48m $\pm$ 0.33& 0.21m $\pm$ 0.21&
19/33&8 \\ \hline H-map \cite{zaenker}& 0.16 $\pm$ 0.09& 0.76m $\pm$
0.49& 0.38m $\pm$ 0.24&14/33&4 \\
\hline \end{tabular}} \caption{Quantitative evaluation with respect to
the average intersection over union~(IoU), center of mass
distance~(CoM-Dist), Hausdorff distance~(H-Dist), as well as the
number of true positive~(TP) and false positive~(FP) object mappings.
Also the standard deviation of each metric is shown.
The results are in comparison to the hypermap framework by Zaenker
\etal \cite{zaenker} where the metrics marked with~a~$\ast$ are significant according to the Wilcoxon sign rank test.
As can be seen our approach outperform the hypermap in all but the FPs.
We explain the higher FP values with the added objects by the object refinement.
with more observations this value will decrease.
} 
\label{ev_res}
\end{table*} 
\begin{figure}
\centering
\subfloat[Handcrafted ground truth\label{office_eval_1_pic:a}]{\includegraphics[height=1.7in]{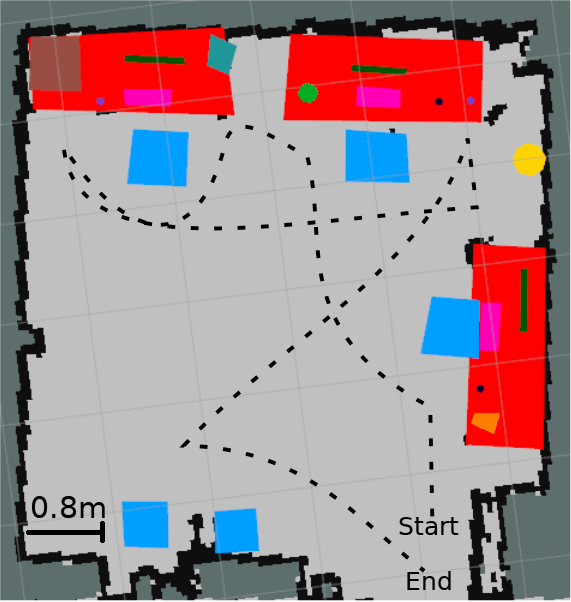}}
\hspace{\fill}
\subfloat[Our approach\label{office_eval_1_pic:b}]{\includegraphics[height=1.7in]{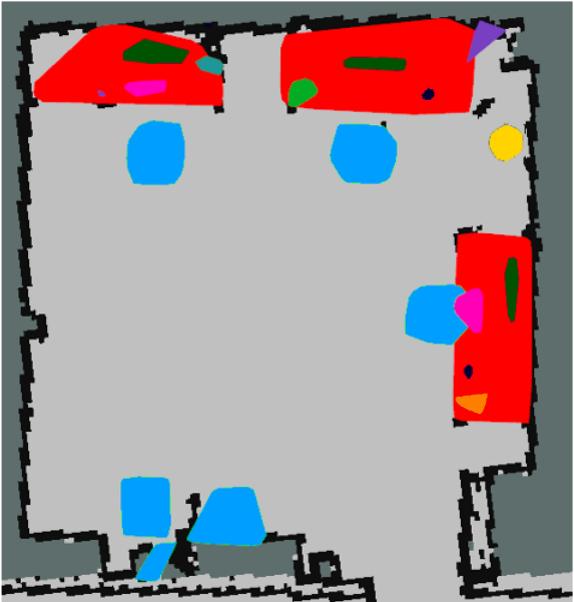}}\\
\subfloat[Hypermap result]{\includegraphics[height=1.7in]{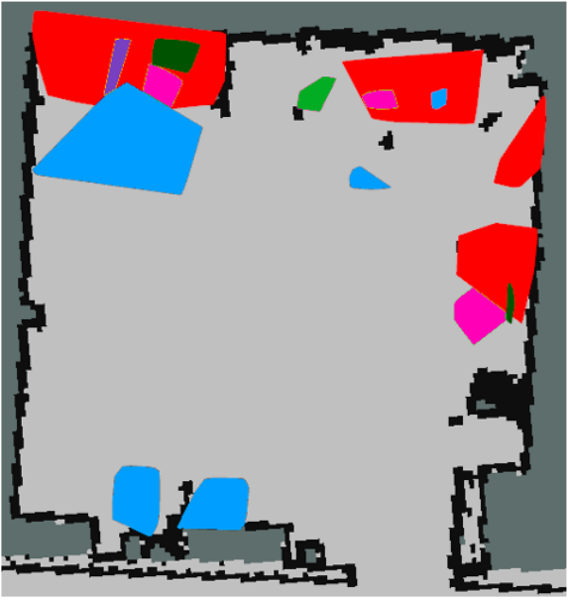}}
\hspace{\fill}
\subfloat[Legend]{\includegraphics[height=1.7in]{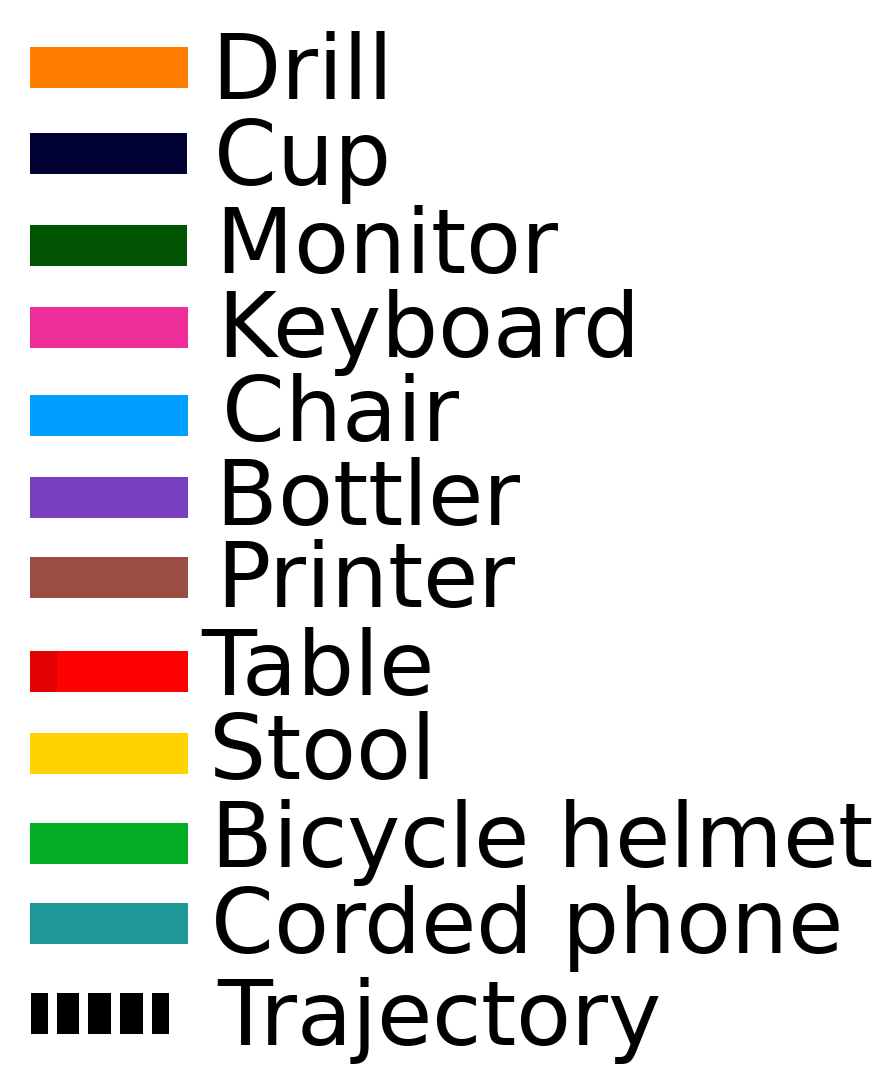}}
\caption{(a) Visualization of the handcrafted ground truth map, (b)
  result of the presented approach, and (c) result of the approach by
  Zaenker~\etal\cite{zaenker}. Each object is represented by a polygon, colored corresponding to the object class shown in (d). While our approach achieves a result close to the ground truth, the hypermap is just a rough approximation.}
\label{office_eval_1_pic}
\end{figure}
To show the performance of our system, we carried out four experiments in three different real-world environments.
For the first three experiments, we used the Toyota HSR~\cite{hsr} and
collected RGB-D data with an Asus Xtion Pro mounted at a height of $1.35m$ and in an angle of~$36^\circ$ from the horizontal facing down.
The last experiment was carried out with the Fraunhofer Care-O-Bot-4~\cite{careobot}, with an Asus Xtion Pro mounted at a height of $1.02m$ horizontal to the floor. 
We used a resolution of $640$x$480$ pixels for the cameras.

The first environment is an office environment of $29 m^2$.
To evaluate the update capability of our system, we split the experiment in the office environment into two parts.
In the second part of the experiment, some objects in the room were displaced, exchanged, removed, or newly added. With $60 m^2$
the second environment is twice as large and consists of a kitchen as well as a sofa corner as shown in Fig \ref{cover}.
The third environment is a coffee room of $35 m^2$ in the Aalto University.
This environment was much more cluttered and therefore more challenging.

We evaluated our approach on the four sequences in comparison to the hypermap (H-Map) framework~\cite{zaenker}.
For each experiment and each approach, we computed the average of each of the following metrics:
\begin{description}[font=\normalfont\itshape] 
\item[Intersection over union (IoU):]
The IoU shows the general similarity of two shapes. For further
evaluations, we assign to
each mapped object  the ground truth object with the highest IoU.
\item[Hausdorff distance (H-dist):] 
The H-dist is the maximum distance in a set of point pairs and indicates the size inaccuracy of the estimated 2D shape
compared to the ground truth.
\item[Distance between the 2D center of mass (CoM):] This metric gives
  the distance between the CoM of two shapes and indicates the
  displacement of the observed object.
\item[True positives (TP) and false positives (FP):] These metrics
  show how many objects are correctly or falsely mapped during one
  experiment. An object mapping is counted as TP if the IoU to the
  groud truth was greater than 0.5 and as FP if there is no
  corresponding groud truth object or the IoU is less than 0.5.
\end{description} 

The results in \tabref{ev_res} show that our approach outperforms
the H-Map in all but the number of FPs~(see also the discussion in \secref{strength}).

To see whether the differences between the results of the metrics are significant, we used the one-sided Wilcoxon sign rank test.
The difference of the results marked with a $\ast$ in \tabref{ev_res} are significant with a chosen p-value of 0.05.
By looking at the resulting maps shown in Fig.
\ref{office_eval_1_pic} the difference between our approach and the H-Map is clearly visible.
Especially, the objects on the tabletops and the chairs are better  approximated and close to the ground truth.

\subsection{Evolution of the Object Likelihood}
\begin{figure}
\centering
\includegraphics[scale=0.55]{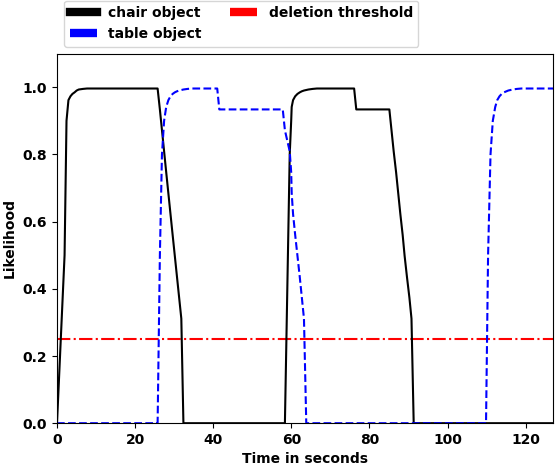}
\caption{Likelihood evolution over time for two different object types.
The likelihood grows as long as the object is detected and decreases otherwise. If the mapped object is not in the robot's field of view the likelihood remains static. The red line shows the threshold of deleting an object from the map.}
\label{plot}
\end{figure}
To show the strength of the object update with likelihood values, we recorded a sequence with changing chair and table appearances over time.
Fig.
\ref{plot} shows the evolution of the likelihood values.
While the robot observes the object, the likelihood grows constantly until it reaches a value of nearly~1.
As long as the robot's camera view does not cover the position of the mapped object, the likelihood is not updated.
If the mapped object is not detected anymore within the FOV, the
likelihood decreases and the object is removed as soon as it reaches
the predefined threshold of 0.25.
In this experiment, we displaced the objects in the scene while they
were not in the robot's FOV.
That is why a small time window exists where both objects are mapped.
The required time to remove an unseen object from the map depends on
the predefined deletion threshold and the maximum number of
observations that are considered for the likelihood calculation.
In the first mapping of the table, the result of an unstable object detection can be seen.
As soon as the robot turns away, it no longer detects the object and therefore the likelihood decreases until the value remains stable when the object is no longer in the FOV.
As this example demonstrates, our system is able to keep the map
updated in case object locations change and can also deal with false detections and dynamic objects.

\subsection{Run Time and Memory Consumption}

\begin{table}
    \centering
\resizebox{.4\textwidth}{!}{%
    \begin{tabular}{|c|c|}
         \hline 
  \rowcolor[gray]{.75} \textbf{Component} & \textbf{Avg time}  \\

  Point cloud transformation& 7 ms  \\
  \hline
  Preprocessing & 31 ms  \\
    \hline
  Euclidean segmentation & 0.01 ms  \\
  \hline
  Data association and object updates $\ast$& 11 ms \\
  \hline
  Likelihood computation and object deletion& 2 ms \\
  \hline
  \hline
Complete average run time per frame & 85 ms \\
  \hline
  Faster R-CNN & 425 ms  \\
  \hline
    \end{tabular}}
    \caption{Average run time per frame of the main components of our
      system and the complete run time for a dataset of the office
      environment. The run time of the component marked with a $\ast$
      is the average computation time for one object. All components
      are included in the 85\,ms.  Note that the output of faster R-CNN
      can be computed parallel to our system. The run time of the
      object detector depends highly on the used network and is not included in the overall run time of our system.}
    \label{time}
\end{table}

To show that our system is capable of online mapping, we also evaluated the run time.
To measure the time of each component, we used the data set of the office environment.
We recorded the experiment over a length of 6.36 minutes and up to 33 mapped objects.
The resulting time, shown in \tabref{time} is the average
computation time per frame over the whole data set. It can be seen that the overall run-time of the office environment is 85\,ms (11.76 Hz).
Most of the time is taken by the outlier removal in the preprocessing
and by each object update, which can vary depending on the number of
mapped objects and the size of the point cloud.
The resulting computation time demonstrates the online mapping capability of
our system and the increased runtime performance in comparison to the
approaches by S{\"u}nderhauf~\etal\cite{meaningful} and and \mbox{Li~\etal\cite{Li}}.

\section{Discussion}
\label{strength}
The experiments show that our system is capable of online semantic
mapping and robust to changes in the environment.
 A videos  of  the  experiments  can  be  found  at \url{https://youtu.be/JoUEW_-VXq0}. However, during the
experiments we encountered a certain amount of false positive
mappings~(cf. \tabref{ev_res}), an example of this can be seen around
the lower two chairs in \figref{office_eval_1_pic:b}, where an
additional shape was falsely added. These mappings can be traced back
to the object refinement step where each cluster is assigned to a new
object. However, with a longer observation time per observed area or a
higher likelihood threshold, the false positives will disappear. This
conclusion can be drawn from our results, as each false positive object
has a likelihood of less than 0.5, which is lower than the initial likelihood.

\section{Conclusion}
In this paper, we presented an approach to online semantic mapping
with modular object representations.
Our approach differs from previous works in the way the created map is represented so that it is not restricted to just one representation but maintains multiple of representations within one framework.
Furthermore, the introduced likelihood calculation of each object is a way to keep the map updated in case objects change their location and achieve robustness in case of false detections.
Our object refinement step ensures that object point assignments can be undone to deal with wrong data association. 
We demonstrated the efficacy of our approach in four real-world
experiments where the results show the increased performance in
comparison to an existing framework with respect to various metrics.
The evaluation of the run time highlights that our system is capable of online mapping while the robot is moving through the environment.
The code of our system can be found at \mbox{\url{https://github.com/NilsDengler/sem_mapping}}.



\bibliographystyle{IEEEtran}
\bibliography{IEEEabrv,references}

\begin{thebibliography}{10}
\providecommand{\url}[1]{#1}
\csname url@rmstyle\endcsname
\providecommand{\newblock}{\relax}
\providecommand{\bibinfo}[2]{#2}
\providecommand\BIBentrySTDinterwordspacing{\spaceskip=0pt\relax}
\providecommand\BIBentryALTinterwordstretchfactor{4}
\providecommand\BIBentryALTinterwordspacing{\spaceskip=\fontdimen2\font plus
\BIBentryALTinterwordstretchfactor\fontdimen3\font minus
  \fontdimen4\font\relax}
\providecommand\BIBforeignlanguage[2]{{%
\expandafter\ifx\csname l@#1\endcsname\relax
\typeout{** WARNING: IEEEtran.bst: No hyphenation pattern has been}%
\typeout{** loaded for the language `#1'. Using the pattern for}%
\typeout{** the default language instead.}%
\else
\language=\csname l@#1\endcsname
\fi
#2}}

\bibitem{slam}
C.~Cadena, L.~Carlone, H.~Carrillo, Y.~Latif, D.~Scaramuzza, J.~Neira, I.~Reid,
  and J.~J. Leonard, ``{Past, Present, and Future of Simultaneous Localization
  And Mapping: Towards the Robust-Perception Age},'' \emph{IEEE Transactions on
  robotics}, vol.~32, no.~6, pp. 1309--1332, 2016.

\bibitem{kimera}
A.~Rosinol, M.~Abate, Y.~Chang, and L.~Carlone, ``{Kimera: an Open-Source
  Library for Real-Time Metric-Semantic Localization and Mapping},'' in
  \emph{Proc.~of the IEEE Intl.~Conf.~on Robotics \& Automation (ICRA)}, 2020,
  pp. 1689--1696.

\bibitem{meaningful}
N.~S{\"u}nderhauf, T.~T. Pham, Y.~Latif, M.~Milford, and I.~Reid, ``{Meaningful
  Maps With Object-Oriented Semantic Mapping},'' in \emph{Proc.~of the IEEE/RSJ
  Intl.~Conf.~on Intelligent Robots and Systems (IROS)}, 2017, pp. 5079--5085.

\bibitem{Li}
W.~Li, J.~Gu, B.~Chen, and J.~Han, ``{Incremental Instance-Oriented 3D Semantic
  Mapping via RGB-D Cameras for Unknown Indoor Scene},'' \emph{Discrete
  Dynamics in Nature and Society}, vol. 2020, 2020.

\bibitem{zaenker}
T.~Zaenker, F.~Verdoja, and V.~Kyrki, ``{Hypermap Mapping Framework and its
  Application to Autonomous Semantic Exploration},'' in \emph{Proc.~of the IEEE
  Conference on Multisensor Fusion and Integration (MFI)}, 2020.

\bibitem{efficient}
Y.~Nakajima and H.~Saito, ``{Efficient Object-Oriented Semantic Mapping With
  Object Detector},'' \emph{IEEE Access}, vol.~7, pp. 3206--3213, 2018.

\bibitem{voxx}
M.~{Grinvald}, F.~{Furrer}, T.~{Novkovic}, J.~J. {Chung}, C.~{Cadena},
  R.~{Siegwart}, and J.~{Nieto}, ``{Volumetric Instance-Aware Semantic Mapping
  and 3D Object Discovery},'' \emph{IEEE Robotics and Automation Letters
  (RA-L)}, vol.~4, no.~3, pp. 3037--3044, July 2019.

\bibitem{regier2020classifying}
P.~Regier, A.~Milioto, C.~Stachniss, and M.~Bennewitz, ``{Classifying Obstacles
  and Exploiting Class Information for Humanoid Navigation Through Cluttered
  Environments},'' \emph{The Int.~Journal of Humanoid Robotics (IJHR)},
  vol.~17, no.~2, pp. 2\,050\,013--1, 2020.

\bibitem{leidner2014object}
D.~Leidner, A.~Dietrich, F.~Schmidt, C.~Borst, and A.~Albu-Sch{\"a}ffer,
  ``{Object-Centered Hybrid Reasoning for Whole-Body Mobile Manipulation},'' in
  \emph{Proc.~of the IEEE Intl.~Conf.~on Robotics \& Automation (ICRA)}, 2014,
  pp. 1828--1835.

\bibitem{he2017mask}
K.~He, G.~Gkioxari, P.~Doll{\'a}r, and R.~Girshick, ``Mask r-cnn,'' in
  \emph{Proc.~of the IEEE Intl.~Conf.~on Computer Vision (ICCV)}, 2017, pp.
  2961--2969.

\bibitem{yolov2}
J.~Redmon, S.~Divvala, R.~Girshick, and A.~Farhadi, ``{You Only Look
  Once:Unified, Real-Time Object Detection},'' in \emph{Proc.~of the IEEE
  Conf.~on Computer Vision and Pattern Recognition (CVPR)}, 2016, pp. 779--788.

\bibitem{rostf}
``{Tensorflow object detection},''
  \url{https://github.com/osrf/tensorflow_object_detector}, last accessed 29.
  June 2020.

\bibitem{ssd}
W.~Liu, D.~Anguelov, D.~Erhan, C.~Szegedy, S.~Reed, C.-Y. Fu, and A.~C. Berg,
  ``{SSD: Single Shot Multibox Detector},'' in \emph{Proc.~of the
  Europ.~Conf.~on Computer Vision (ECCV)}.\hskip 1em plus 0.5em minus
  0.4em\relax Springer, 2016, pp. 21--37.

\bibitem{ren2015faster}
S.~Ren, K.~He, R.~Girshick, and J.~Sun, ``{Faster R-CNN: Towards Real-Time
  Object Detectionwith Region Proposal Networks},'' in \emph{Advances in neural
  information processing systems}, 2015, pp. 91--99.

\bibitem{yolact++}
D.~Bolya, C.~Zhou, F.~Xiao, and Y.~J. Lee, ``Yolact++: Better real-time
  instance segmentation,'' \emph{arXiv preprint arXiv:1912.06218}, 2019.

\bibitem{rosu2020semi}
R.~A. Rosu, J.~Quenzel, and S.~Behnke, ``{Semi-Supervised Semantic Mapping
  through LabelPropagation with Semantic Texture Meshes},''
  \emph{Intl.~Journal~of Computer Vision (IJCV)}, vol. 128, no.~5, pp.
  1220--1238, 2020.

\bibitem{stekovic2020casting}
S.~Stekovic, F.~Fraundorfer, and V.~Lepetit, ``{Casting Geometric Constraints
  in Semantic Segmentation as Semi-Supervised Learning},'' in \emph{Proc.~of
  the IEEE Winter Conference on Applications of Computer Vision (WACV)}, 2020,
  pp. 1854--1863.

\bibitem{Coco}
T.-Y. Lin, M.~Maire, S.~Belongie, J.~Hays, P.~Perona, D.~Ramanan,
  P.~Doll{\'a}r, and C.~L. Zitnick, ``{Microsoft COCO: Common Objects in
  Context},'' in \emph{Proc.~of the Europ.~Conf.~on Computer Vision
  (ECCV)}.\hskip 1em plus 0.5em minus 0.4em\relax Springer, 2014, pp. 740--755.

\bibitem{oid}
I.~Krasin, T.~Duerig, N.~Alldrin, V.~Ferrari, S.~Abu-El-Haija, A.~Kuznetsova,
  H.~Rom, J.~Uijlings, S.~Popov, S.~Kamali, M.~Malloci, J.~Pont-Tuset, A.~Veit,
  S.~Belongie, V.~Gomes, A.~Gupta, C.~Sun, G.~Chechik, D.~Cai, Z.~Feng,
  D.~Narayanan, and K.~Murphy, ``Openimages: A public dataset for large-scale
  multi-label and multi-class image classification.'' \emph{Dataset available
  from https://storage.googleapis.com/openimages/web/index.html}, 2017.

\bibitem{fischler1981random}
M.~A. Fischler and R.~C. Bolles, ``{Random Sample Consensus: A Paradigm for
  Model Fitting with Apphcatlons to Image Analysis and Automated
  Cartography},'' \emph{Communications of the ACM}, vol.~24, no.~6, pp.
  381--395, 1981.

\bibitem{Rusu_ICRA2011_PCL}
R.~B. Rusu and S.~Cousins, ``{3D is here: Point Cloud Library (PCL)},'' in
  \emph{Proc.~of the IEEE Intl.~Conf.~on Robotics \& Automation (ICRA)},
  Shanghai, China, May 9-13 2011.

\bibitem{calipers}
H.~Freeman and R.~Shapira, ``{Determining the Minimum-Area Encasing Rectangle
  for an Arbitrary Closed Curve},'' \emph{Communications of the ACM}, vol.~18,
  no.~7, pp. 409--413, 1975.

\bibitem{rtree}
A.~Guttman, ``{R-Trees: A Dynamic Index Structure for Spatial Searching},'' in
  \emph{Proc.~of the ACM SIGMOD international conference on Management of data
  (MOD)}, 1984, pp. 47--57.

\bibitem{survival}
D.~M. Rosen, J.~Mason, and J.~J. Leonard, ``Towards lifelong feature-based
  mapping in semi-static environments,'' in \emph{2016 IEEE International
  Conference on Robotics and Automation (ICRA)}.\hskip 1em plus 0.5em minus
  0.4em\relax IEEE, 2016, pp. 1063--1070.

\bibitem{fox1999monte}
D.~Fox, W.~Burgard, F.~Dellaert, and S.~Thrun, ``{Monte Carlo Localization:
  Efficient Position Estimation for Mobile Robots},'' \emph{aaai}, vol. 1999,
  no. 343-349, pp. 2--2, 1999.

\bibitem{gmapping}
G.~Grisetti, C.~Stachniss, and W.~Burgard, ``{Improved Techniques for Grid
  Mapping with Rao-Blackwellized Particle Filters},'' \emph{IEEE transactions
  on Robotics}, vol.~23, no.~1, pp. 34--46, 2007.

\bibitem{hsr}
T.~Yamamoto, T.~Nishino, H.~Kajima, M.~Ohta, and K.~Ikeda, ``{Human Support
  Robot (HSR)},'' in \emph{ACM SIGGRAPH emerging technologies}, 2018, pp. 1--2.

\bibitem{careobot}
R.~Kittmann, T.~Fr{\"o}hlich, J.~Sch{\"a}fer, U.~Reiser, F.~Wei{\ss}hardt, and
  A.~Haug, ``Let me introduce myself: I am care-o-bot 4, a gentleman robot,''
  \emph{Proc.~of Mensch und computer (MUC)}, 2015.

\end{thebibliography}

\end{document}